\documentclass[letterpaper, 10 pt, conference]{ieeeconf}  

\IEEEoverridecommandlockouts                              

\usepackage{lipsum} 
\usepackage{epsfig}
\usepackage{float}
\usepackage{bbold}

\usepackage{amsmath}

\title{\LARGE \bf
Domain Randomization for Sim2real Transfer of Automatically Generated Grasping Datasets
}

\begin{document}

\author{Johann Huber$^{*\dagger}$, François Hélénon$^{*\dagger}$, Hippolyte Watrelot$^{\dagger}$, Faïz Ben Amar$^{\dagger}$ and Stéphane Doncieux$^{\dagger}$ %
\thanks{$^{*}$\hspace{1mm} Equal contribution}
\thanks{$^{\dagger}$\hspace{1mm} Sorbonne Université, CNRS, Institut des Systèmes Intelligents et de Robotique (ISIR), Paris, 75005, France}
\thanks{Corresponding author: johann.huber@isir.upmc.fr}
}

\maketitle
\thispagestyle{empty}
\pagestyle{empty}





\begin{abstract}

Robotic grasping refers to making a robotic system pick an object by applying forces and torques on its surface. Many recent studies use data-driven approaches to address grasping, but the sparse reward nature of this task made the learning process challenging to bootstrap. To avoid constraining the operational space, an increasing number of works propose grasping datasets to learn from. But most of them are limited to simulations. The present paper investigates how automatically generated grasps can be exploited in the real world. More than 7000 reach-and-grasp trajectories have been generated with Quality-Diversity (QD) methods on 3 different arms and grippers, including parallel fingers and a dexterous hand, and tested in the real world. Conducted analysis on the collected measure shows correlations between several Domain Randomization-based quality criteria and sim-to-real transferability. Key challenges regarding the reality gap for grasping have been identified, stressing matters on which researchers on grasping should focus in the future. A QD approach has finally been proposed for making grasps more robust to domain randomization, resulting in a transfer ratio of $84\%$ on the Franka Research 3 arm.

\end{abstract}





\section{INTRODUCTION}
Grasping refers to making a robot solidarize its end effector with an object by applying forces and torques on its surface. This skill is of great interest in robotics as it is a prerequisite for many manipulation tasks \cite{hodson2018gripping}. The former analytical-based methods \cite{nguyen1988constructing} have slowly been replaced by data-based approaches \cite{zhang2022robotic}. But the hard exploration nature of grasping prevents the learning process bootstrapping, as most of the attempted grasp from a randomly initialized policy result in a null reward \cite{huber2023quality}. Many works make the problem tractable through the use of imitation learning \cite{qin2022from}, parallel grippers \cite{fang2020graspnet}, and top-down movements \cite{yang2023pave}. But these approaches and design choices constrain the operational space, limiting the adaptation capabilities of the learned policies.

The AI breakthroughs of the last decade show that getting the most out of data-driven approaches requires large datasets \cite{deng2009imagenet}\cite{ray2023chatgpt}. Similarly, optimizing robust grasping policies demands large sets of diverse demonstrations to learn from. First works on grasping datasets involve analytics criteria to label data \cite{godlfeder2009columbia}\cite{mahler2017dexnet2}\cite{fang2020graspnet}. Still, these methods are gripper-specific and do not always result in successful grasps on the real robot. An increasing number of works rely on simulation to automatically generate grasping datasets \cite{depierre2018jacquard}\cite{eppner2021acronym}\cite{huber2023quality}. But predicting the probability for these grasps to successfully transfer is an open problem.

\begin{figure}[t]
  \centering
  \includegraphics[scale=0.18]{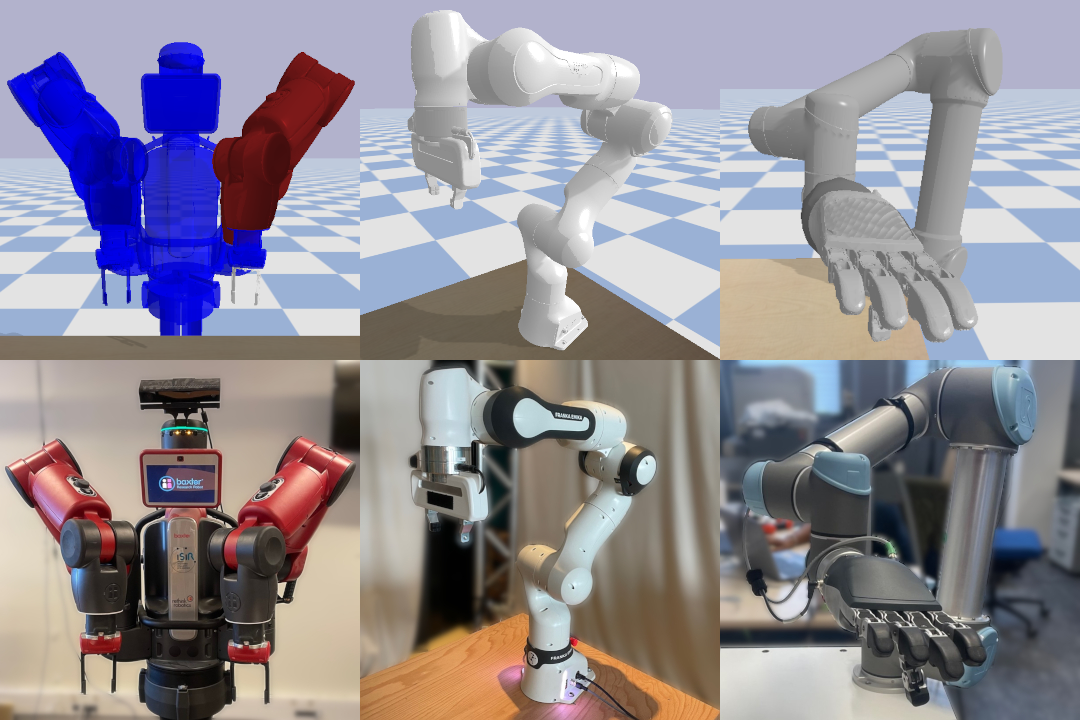}
  \caption{\textbf{Robotic platforms overview.} (left) A Baxter robot with a parallel gripper; (center) a Franka Emika Panda with a parallel gripper; (right) a UR5 robot with a SIH Schunk 5-DoF hand.}
  \label{fig:robotic_plateforms_overview}
\end{figure}

\begin{figure}[t]
  \centering
  \includegraphics[width=0.5\columnwidth]{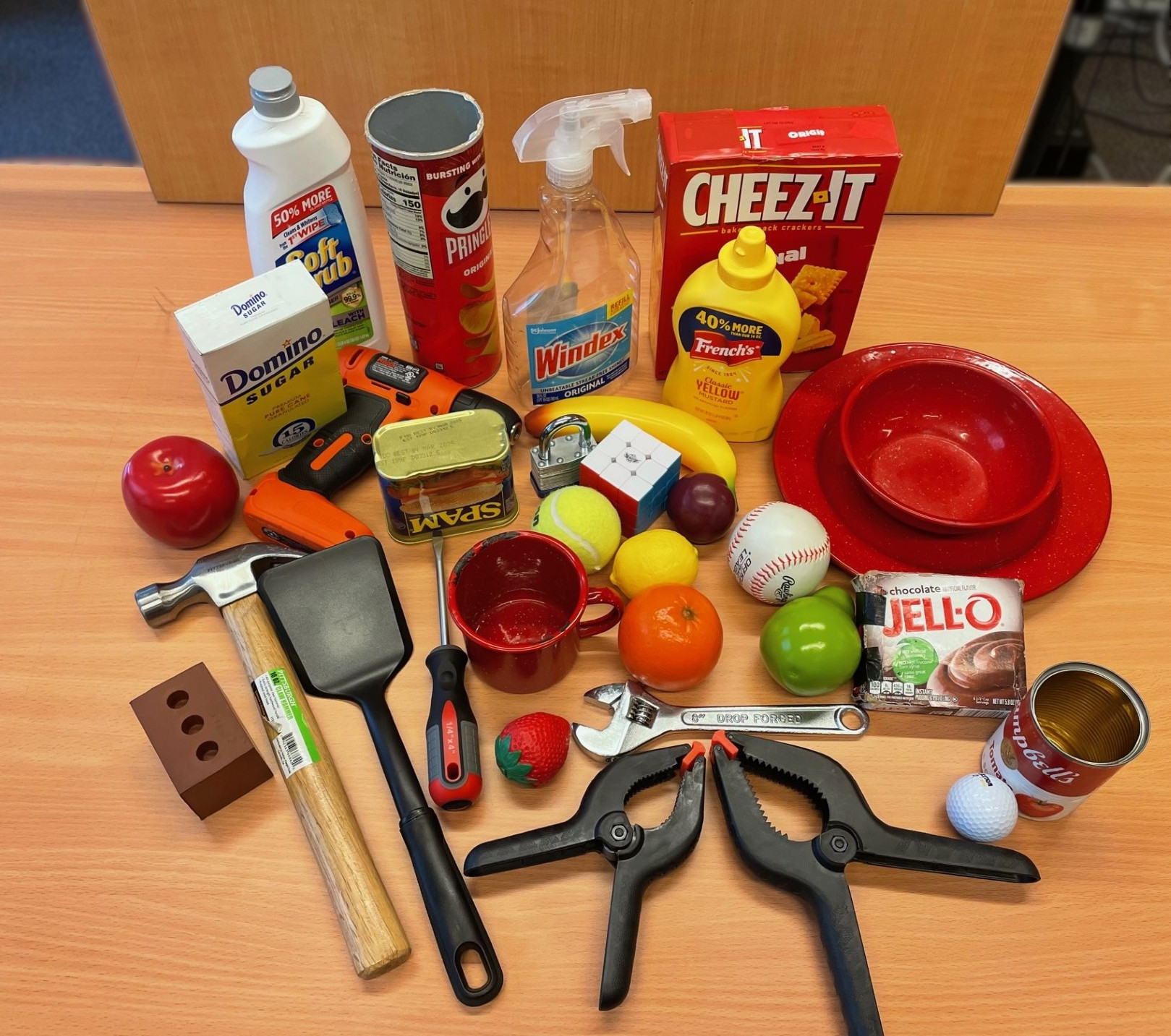}
  \caption{\textbf{Targeted objects overview.} More than 7000 reach-and-grasps trajectories have been deployed on the following 32 YCB objects \cite{calli2015benchmarking} to study the sim2real transferability of automatically generated grasps.}
  \label{fig:ycb_objects_used}
\end{figure}

Domain Randomization (DR) has shown promising results for sim2real transfer of data-driven approaches \cite{akkaya2019solving}\cite{muratore2022robot}. To our knowledge, no systematic study has been conducted to investigate to what extent DR is correlated to sim2real transferability for the generation of grasps. Such an analysis would be of great interest: 1) it would give quantitative results in favor of the usage of DR for grasps generation; 2) a correlation would show that optimizing DR-based quality criteria should result in more transferable grasps. The present study aims to fill this gap by \textbf{extensively studying the relation between the transferability of automatically generated grasps and several DR-based quality criteria.}

Most of the works on grasping datasets are limited to a single kind of gripper. However, the simulation-based grasp generators can easily be applied to different grippers. This idea has been leveraged in the field of Quality-Diversity (QD) algorithms to generate large sets of grasps on different robotic hands \cite{huber2023quality}. The present paper \textbf{investigates the exploitability of QD-generated grasps by testing thousands of solutions in the real world.}

These work contributions are the following:

\begin{itemize}
    \item We have generated and deployed more than 7000 grasping trajectories on 3 robotic platforms and 32 objects, measuring their sim2real transferability and identifying the main causes of the reality gap;
    \item We show a significant correlation between 4  investigated domain randomization-based quality criteria and the sim2real transferability;
    \item We proposed a QD approach for making generated grasps more robust to sim2real transfer, reaching a transfer ratio of $84\%$ on a Franka Research 3 robot.
    \end{itemize}

A video presenting the conducted experiments can be found online\footnote{Attached to the paper and released soon.}. The code is available on Github\footnote{https://github.com/Johann-Huber/qd-grasp}. We believe this work to be a step toward the exploitation of automatically generated grasping datasets.

\section{RELATED WORKS}
\label{sec:2_related_works}


\textbf{\textit{Datasets for grasping in robotics.}}
The interest in data-driven approaches pushes the field to produce many datasets for grasping in robotics. The data can be manually annotated \cite{jiang2011efficient} or acquired through self-supervision \cite{dasari2019robonet}. Both those approaches are very time and cost-expensive, restricting them to constrained operational spaces. Data can be automatically labeled using analytics criteria \cite{godlfeder2009columbia}\cite{fang2020graspnet}, but it is not always successful into the real world \cite{mahler2017dexnet2}\cite{lundell2021multi}\cite{siddiqui2021grasp}. Recent works rely on a physics engine to simulate the grasp interaction \cite{depierre2018jacquard}\cite{eppner2021acronym}\cite{turpin2023fast}, but those works focus on a specific type of gripper, and do not imply extensive sim-to-real experiments. Recently, Huber et al. \cite{huber2023quality} demonstrated that Quality-Diversity (QD) methods can generate datasets of diverse and high-performing reach-and-grasp trajectories for different types of grippers. But these results are limited to simulations. The present paper investigates how such automatically produced grasping demonstrations can be exploited in the real world by identifying quality criteria correlated with sim2real transfer.

\textbf{\textit{Bridging the reality gap.}}
The difference between simulation and reality (\textit{reality gap}) is well-known as a major challenge in robotics \cite{salvato2021crossing}. Many approaches have been proposed to face it. In evolutionary robotics, the \textit{transferability approach} has been introduced to bridge the reality gap by leveraging results from real experiments \cite{koos2012transferability}. Most of the commonly used approaches optimize transferable solutions by collecting real-world data during the learning process  \cite{jimenez2021model}\cite{jiang2021simgan}\cite{koos2012transferability}, but real robots experiments are very time and cost expensive \cite{ibarz2021train}. 
Domain randomization (DR) consists of adding noise into the simulation's signals (e.g. observations, actions, dynamics) to make the generated solutions more robust to sim-to-real transfer \cite{muratore2022robot}. The significant results obtained with DR \cite{akkaya2019solving} incite the authors to use it for 2 purposes: identify transferable grasps and robustify them. 

\textbf{\textit{Predicting sim-to-real transferability.}}
Most of grasping works rely on analytics criteria \cite{nguyen1988constructing} to identify a \textit{force closure} \cite{miller2004graspit} \cite{fang2020graspnet}. Even though it guarantees a successful grasp in the simulated scene under some friction conditions, the reality gap leads to many failing grasps \cite{mahler2017dexnet2}\cite{lundell2021multi}\cite{siddiqui2021grasp}. 

Several works quantify the difference between simulation and reality. Koos et al. \cite{koos2012transferability} measure the sim-to-real error to guide the optimization process toward robust solutions. Kadian et al. \cite{kadian2020sim2real} propose a metric associated with the length of the path traveled by the robot and estimate the correlation between measures in simulation and in reality to adjust their model. Collins et al. \cite{collins2019quantifying} measure the differences between the output signal on a simulated robot and on its real equivalent. Recently, Enayati et al. \cite{enayati2023facilitating} studied the correlation between the instrinsic stochasticity of real-time simulation and the noise of real-world dynamics. These works rely on a continuous metric, either because the navigation tasks are suited to do so \cite{koos2012transferability}\cite{kadian2020sim2real} or by measuring the error on low-level output signal \cite{collins2019quantifying}\cite{enayati2023facilitating}. 

To label grasping data, what matters is the grasping result: measuring error on the actuators' signal might not be informative, as a large error on some part of a trajectory will not prevent the grasp. Contrary to these related works, we decided to focus on the binary outcome of the grasping attempt. Another notable difference is that the experiments have been conducted on 3 different robot-gripper setups while all of these work are usually limited to a single setup with parallel fingers. It shows the capabilities of QD methods to generate grasps for multiple robotic platforms and demonstrates that similar results on sim2real prediction can be obtained on common robotic manipulators.


\section{METHOD}




\subsection{Notations}

Let $S_\tau \subseteq \mathbb{R}^n$ be the \textit{space of grasping trajectories}. Let $\tau\in S_\tau$ be a \textit{trajectory} defined as a sequence of joint states $X_i$ and torques $\Gamma_j$ for $ i \in \left\{0, ...,  T \right\}$ and $ j \in \left\{0, ...,  T-1 \right\}$, with $T\in\mathbb{N}^{+*}$ being the length of the sequence (\textit{episode}). A grasping simulated scene is submitted to a \textit{transition function} $\rho:(X_k,\Gamma_k)\mapsto X_{k+1}$. In this paper, grasps are generated under a deterministic $\rho$. Let $f_c(\tau):S_\tau\rightarrow  \left\{0, 1 \right\}$ be the \textit{grasping success criteria} (similarly defined as \cite{huber2023quality}). In practice, the transition function $\rho$ is defined by the physics engine of the simulator that approximates real-world physics. Let $\zeta_r, \zeta_s \in  \left [ 0,1 \right ]$ respectively be the \textit{rolling} and \textit{spinning friction coefficients}. A targeted object is initialized at a fixed state $X_0^{obj}$.


\subsection{QD for grasping}

Quality-Diversity (QD) methods are evolutionary algorithms that optimize both a quality criterion (called the \textit{fitness}) and the diversity of solutions for a given problem \cite{cully2022quality}. These algorithms can be used to generate large grasping datasets \cite{huber2023quality}. The present work leverages QD methods to generate diverse grasps and study their transferability. To limit the impact of the grasp generator on the collected results, the deployed trajectories have been produced with 3 methods: NSMBS \cite{morel2022automatic} is the first QD method that addresses grasping; DC-NSMBS \cite{huber2023diversity} has later been introduced to tackle NSMBS limited diversity issue; ME-scs \cite{huber2023quality} is a variant of MAP-Elites \cite{mouret2015illuminating} that has recently been shown as the most efficient QD method to generate a large set of diverse grasps. While ME-scs optimizes a real-valued quality criterion, NSMBS variants only consider the binary outcome of the grasp – resulting into different solutions. \textbf{By comparing results on each of these methods, this work aims to avoid drawing conclusions that might overfit a specific way to generate the grasps.}


\subsection{Domain Randomization as quality criteria}

Many variants of DR can be found in the literature, such that successfully applying these methods is subject to many questions \cite{muratore2022robot}. DR adds perturbations to the optimization process to make the generated solutions more robust to uncertainties. Thus, \textbf{it is critical to make these perturbations fit the uncertainty imposed by the sim2real transfer on grasping.} Designing the perturbation scale is also crucial: too small perturbations will not make the solutions more robust to sim2real transfer, while too large perturbations can prevent the learning process. We have thus decided to focus on sources of uncertainty that are likely to have a strong impact on sim2real transfer for grasping.

\textit{\textbf{Uncertainty on perception.}} The present work focuses on vision-based grasping approaches, as it is the most common setup in the literature \cite{newbury2023review}. But designing reliable vision systems for robotic grasping is a challenging task \cite{bai2020survey}: problems to address include the sensor noises and the complex input distribution that image processing methods must catch. The perturbation of the object 6-DoF pose estimation is therefore considered among the tested variants (\textit{Object state DR}). The associated fitness is defined as:
\begin{equation*}
    f^{OSDR}(\tau)=\frac{\sum_{i=1}^{N_{OSDR}}f_c^{OSDR}(\tau_i)}{N_{OSDR}}
\end{equation*}
with $N_{OSDR}\in\mathbb{N}^{+*}$, and $f^{OSDR}:S_\tau\rightarrow  \left\{0, 1 \right\}$ being the success criterion evaluating $\tau$ on an object initially positioned at the perturbated state $\tilde{X}_0^{obj}=X_0^{obj}+\delta_0$, $ \delta_0 \sim \mathcal{N}(0,\sigma_0)$.

\textit{\textbf{Uncertainty on joint states.}} Real robots are submitted to stochasticity due to variations of the environment's conditions or wear-and-tear \cite{asada2011special}, causing reproducibility issues \cite{ibarz2021train}. The robustness to the variance of joint states should thus be informative regarding the transferability of a grasping trajectory (\textit{Joint states DR}). The associated fitness is:
\begin{equation*}
    f^{JSDR}(\tau)=\frac{\sum_{i=1}^{N_{JSDR}}f_c^{JSDR}(\tau_i)}{N_{JSDR}}
\end{equation*}
with $N_{JSDR}\in\mathbb{N}^{+*}$, where $f^{JSDR}(\tau):S_\tau\rightarrow  \left\{0, 1 \right\}$ evaluates $\tau$ under a perturbated transition function $\tilde{\rho}(X,\Gamma)=\rho(X,\Gamma)+\delta_j$, $ \delta_j \sim \mathcal{N}(0, \sigma_j)$.
    
\textit{\textbf{Uncertainty on dynamics.}} Simulators for learning in robotics do not perfectly reproduce the real-world dynamics \cite{collins2021review}. DR on simulator dynamics have already shown promising results \cite{mordotach2015ensemble}\cite{antonova2017reinforcement}. Robustness to perturbation of simulator dynamics is the third investigated criterion (\textit{Frictions DR}). The associated fitness is:
\begin{equation*}
    f^{FDR}(\tau)=\frac{\sum_{i=1}^{N_{FDR}}f_c^{FDR}(\tau_i)}{N_{FDR}}
\end{equation*}
with $N_{FDR}\in\mathbb{N}^{+*}$, where $f^{FDR}(\tau):S_\tau\rightarrow \left\{0, 1 \right\}$ evaluates $\tau$ under perturbated rolling and spinning frictions coefficients respectively noted $\tilde{\zeta_s}$ and $\tilde{\zeta_r}$. These coefficients are defined as $\tilde{\zeta_s} \sim \mathcal{U}(\zeta_s^{min}, \zeta_s^{max})$, and $\tilde{\zeta_r} \sim \mathcal{U}(\zeta_r^{min}, \zeta_r^{max})$, with $\zeta_s^{min}, \zeta_s^{max} \in \left [ 0,1 \right ]$ and $\zeta_r^{min}, \zeta_r^{max} \in \left [ 0,1 \right ]$ being the extrema values of spinning and rolling friction coefficients.

\textit{\textbf{Combined uncertainties.}} While it is possible to investigate each of the abovementioned uncertainties separately in simulation, these issues must be addressed simultaneously when working with real robots. The combination of each of these DR methods is the fourth and the last studied variants (\textit{Mixture DR}). The associated fitness is:
\begin{equation*}
    f^{MDR}(\tau)=\frac{\sum_{i=1}^{N_{MDR}}f_c^{MDR}(\tau_i)}{N_{MDR}}
\end{equation*}
with $N_{MDR}\in\mathbb{N}^{+*}$, where $f^{MDR}(\tau):S_\tau\rightarrow  \left\{0, 1 \right\}$ evaluates $\tau$ under a perturbated object initial state $\tilde{X}_0^{obj}$, transition function $\tilde{\rho}$, and friction coefficients $\tilde{\zeta_s}$ and $\tilde{\zeta_r}$.



\begin{figure}[t]
  \centering
  \includegraphics[scale=1.0]{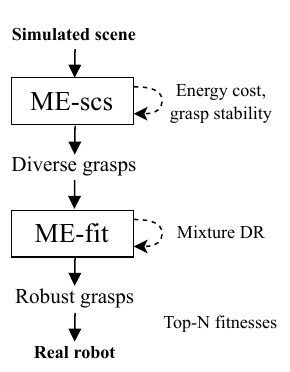}
  \caption{\textbf{TR-ME overall approach for robustifying grasps.}}
  \label{fig:tr_me_principle}
\end{figure}


\subsection{Robustifying grasps}

The promising results obtained on the correlation between DR variants and sim2real transfer lead to the idea of \textbf{making the generated grasps more transferable by optimizing robustness to DR}. As the Mixture DR is likely to mimic the real-world noise more closely, we proposed a QD method to generate diverse and transferable grasps: \textit{Transferability Robustified MAP-Elites – TR-ME} (see Fig. \ref{fig:tr_me_principle}). Similarly to QD methods applied to grasping than can be found in the literature \cite{morel2022automatic}\cite{huber2023diversity}\cite{huber2023quality}, a simulated scene of a robot, a table and an object to grasp is first established. Grasping solutions are generated by exploring the interactions between the robot and a targeted object through an iterative selection and mutation process. A first MAP-Elites \cite{mouret2015illuminating} algorithm is applied with a success-greedy selection process (ME-scs), as it appears to be the most efficient method to generate a large set of diverse grasps \cite{huber2023quality}. In this step, the energy consumption and the grasp stability are optimized – so as diversity. The obtained repertoire of successful solutions is then used to bootstrap a fitness-greedy MAP-Elites (ME-fit), defined as standard MAP-Elites that select the best-performing solutions in priority to guide the optimization process. In this second step, the fitness optimized is the robustness to Mixture DR. \textbf{The outcome is a repertoire of diverse solutions optimized to ensure the grasp success under a large variance of perturbations on the perception, the control, or the dynamics.}


\section{EXPERIMENTS}

\textit{\textbf{Robotic platforms.}} The empirical study conducted in this paper requires a variety of robotic platforms (Fig. \ref{fig:robotic_plateforms_overview}): a Baxter robot with a parallel gripper, a Franka Research 3 (FR3) arm with the standard Franka gripper, and an UR5 robotic arm with a Schunk SIH hand. The experiments thus involve various platforms with diverse robotic arms (6 and 7-DoF) and grippers (2-fingers and 5-DoF dexterous hand). Note that the Schunk hand is controlled with predefined synergies similarly to \cite{helenon2023learning}.

\textit{\textbf{Targeted objects.}} 32 of the YCB objects \cite{calli2015benchmarking} are targeted for the experiment (Fig. \ref{fig:ycb_objects_used}), including small-sized (\textit{padlock}, \textit{strawberry}, \textit{flat screwdriver}, \textit{golf ball}, \textit{plum},  \textit{lemon}), medium-sized (\textit{bowl}, \textit{banana}, \textit{rubiks cube}, \textit{foam brick}, \textit{large clamp}, \textit{mug}, \textit{orange},  \textit{tomato soup can},  \textit{baseball}, \textit{pear},  \textit{tennis ball}, \textit{potted meat can}, \textit{pudding box}, \textit{apple}, \textit{sugar box}, \textit{mustard bottle}) and large-sized (\textit{power drill}, \textit{adjustable wrench},  \textit{extra large clamp},  \textit{hammer}, \textit{bleach cleanser},  \textit{cracker box}, \textit{windex bottle}, \textit{spatula},  \textit{plate}, \textit{chips can}) objects.

\textit{\textbf{Simulation.}} Automatic generation of grasp requires a simulator to experiment in. Following previous QD works for grasping \cite{huber2023quality}, Pybullet \cite{coumans2016pybullet} has been used for the generation of the grasps. This simulator has been shown well aligned with the real world \cite{collins2019quantifying} while assuring many of the expected properties for grasp generation, including high computational speed and short cold restart \cite{mouret2017reality}.

\textit{\textbf{Sim2real deployment of grasping trajectories.}} Thousands of reach-and-grasp trajectories have been produced with 3 different QD methods: \textit{NSMBS} \cite{morel2022automatic}, \textit{DC-NSMBS} \cite{huber2023diversity}, and \textit{ME-scs} \cite{huber2023quality}. The real scene is then built to match the simulation. Objects and robots are positioned as accurately as possible with different devices: Optitrack sensors for Baxter, point cloud matching on a RealSense D455 for the UR5-SIH, and using the forward kinematics of the robot's gripper for the FR3 arm. Trajectories are deployed in an open-loop manner following the simulation. Note that another work investigates how such grasps can be adapted to any other states through vision-based adjustment \cite{helenon2023learning}. Objects are manually reset between each deployment. A grasp is considered successful if the object is not falling after applying an external perturbation to the end effector.

\textit{\textbf{Robustification evaluation.}} Grasping sets that have been previously generated with ME-scs have then been used to bootstrap a ME-fit algorithm, as described in Fig. \ref{fig:tr_me_principle}. ME-scs and ME-fit algorithms are extensively detailed in \cite{huber2023quality}. The generated grasp that maximizes robustness to Mixture DR is then deployed into the real world. The evaluation has been conducted on the FR3 arm, as it is one of the robots on which fewer trajectories have been deployed for the sim2real systematic studies.

\textbf{Hyperparameters.} The parameters used to generate the grasps match those described in each of the QD grasping papers \cite{morel2022automatic}\cite{huber2023diversity}\cite{huber2023quality}. Grasping repertoires are obtained after 400k evaluations of robot-object interaction, and then deployed into the real world. Hyperparameters associated with DR fitnesses are the following: $\zeta_s=0.1$, $\zeta_s^{min}=0.1$, $\zeta_s^{max}=0.4$, $\zeta_r=0.01$, $\zeta_r^{min}=0.01$, $\zeta_r^{max}=0.04$ (tuned by hand to make contact move from sticking to slippery),  $N_{OSDR}=N_{JSDR}=N_{FDR}=N_{MDR}=100$, $\sigma_0=0.005 m$ (error measured on our vision-based 6DoF pose estimation module \cite{helenon2023learning}), $\sigma_j=0.002$ (measured noise on joint poses).

\begin{figure}[t]
  \centering
  \includegraphics[width=0.8\columnwidth]{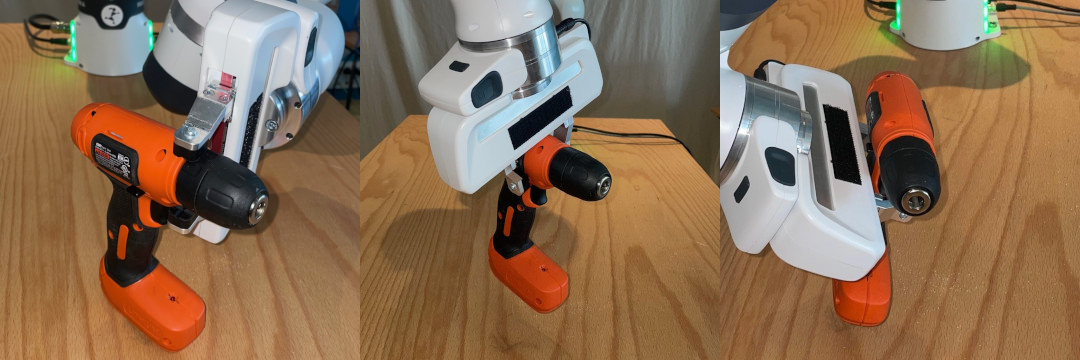}
  \caption{Example of automatically generated diverse grasps that successfully transferred into the real world.}
  \label{fig:diverse_grasps_s2r_franka}
\end{figure}

\begin{table}[t]
\centering
\begin{tabular}{ ||c || c | c | c  ||}
\hline
 & \multicolumn{3}{c||}{\textbf{methods}}  \\
\hline
 & NSMBS & DC-NSMBS & ME-scs  \\
\hline
 $\eta^{sim2real}$  & 0.31 & 0.33 & \textbf{0.41} \\
\hline
\hline
 & \multicolumn{3}{c||}{\textbf{robots}}  \\
\hline
 & Baxter & UR5-SIH & FR3 \\
\hline
 $\eta^{sim2real}$  & 0.33 & 0.38 & \textbf{0.44} \\
\hline
\hline
 & \multicolumn{3}{c||}{\textbf{objects}}  \\
\hline
 & small & medium & large \\
\hline
 $\eta^{sim2real}$  & 0.38 & \textbf{0.44} & 0.29  \\
\hline

\end{tabular}
\caption{\textbf{Measured sim2real transfer ratios for different setups.}}
\label{table:s2r_deploy_results}
\end{table}



\begin{figure*}[thpb]
  \centering
  \includegraphics[scale=1.0]{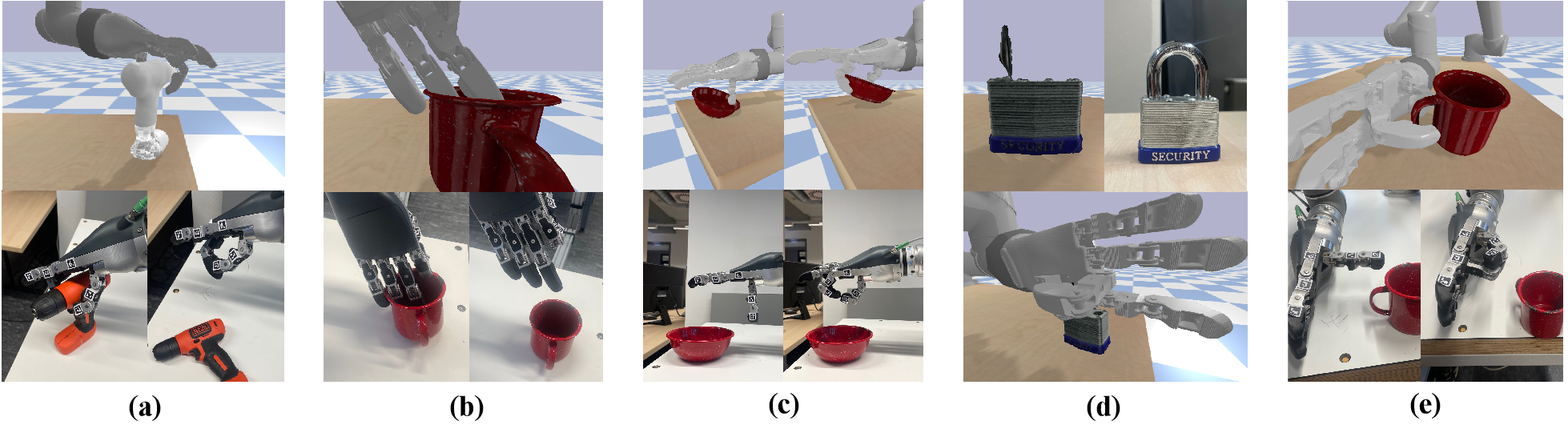}
  \caption{\textbf{Reality gap identified issues.} The most common source of sim-to-real failure are the following issues: \textbf{coarse approximation of frictions (a), contacts (b), or mass distribution (c)} – all top images show successful grasps, that fail in reality; \textbf{errors in the simulated models (d)} – the below-shown solution grasp a part of the object that does not exist in reality; or simply \textbf{fragile grasps (e)} resulting from the exploratory power of QD methods.}
  \label{fig:reality_gap_issues}
\end{figure*}



\begin{figure}[t]
  \centering
  \includegraphics[width=\columnwidth]{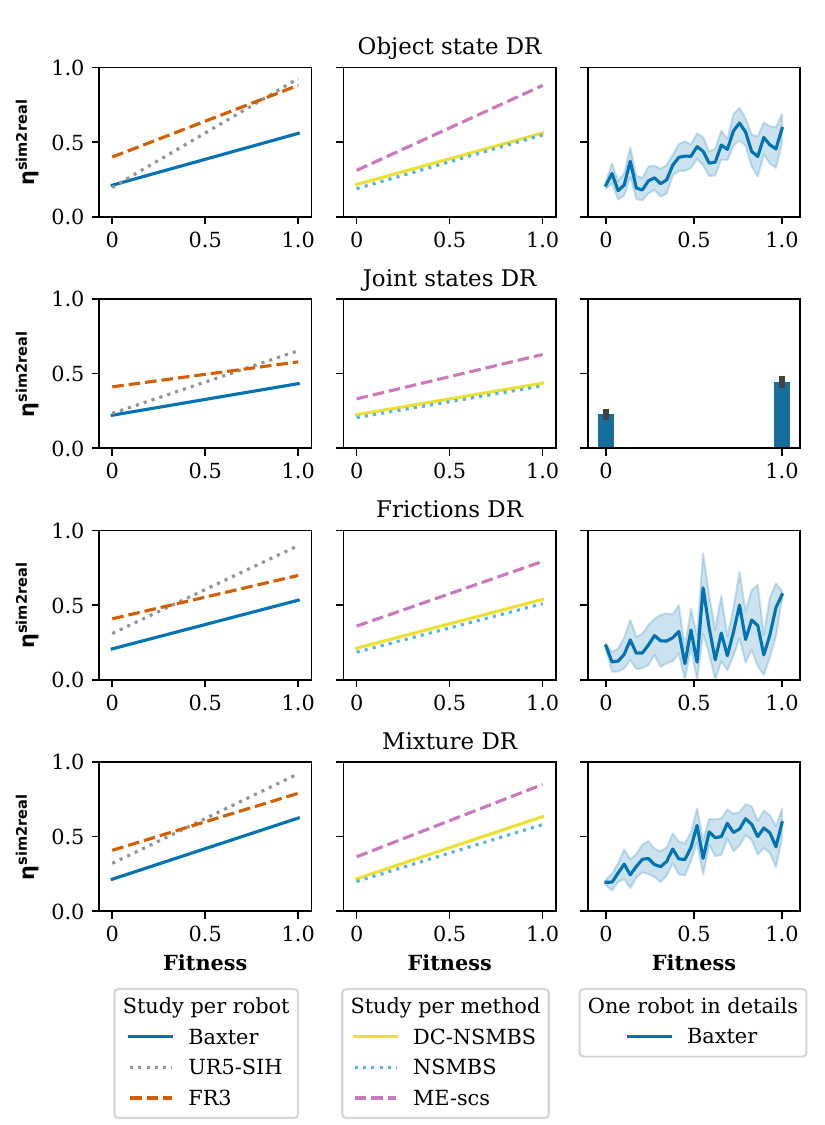}
  \caption{\textbf{Correlation between Domain Randomization (DR) fitnesses and the sim2real transferability ratio ($\eta^{sim2real}$).} There is a correlation for each investigated DR method regardless of the algorithm used to generate the grasps for all the considered robots ($p \leq 0.01$).}
  \label{fig:dr_fit_sim2real_transfert_ratio_results}
\end{figure}


\section{RESULTS AND DISCUSSION}


\textit{\textbf{Trajectories deployment.}}
Table \ref{table:s2r_deploy_results} gives an overview of the obtained results. Overall, $7183$ trajectories have been deployed of the considered platforms ($6815$ on the Baxter, $177$ on the FR3, and $191$ on the UR5 arm with SIH hand). Among the automatically generated trajectories, a ratio of $\eta_{s}=34\%$ deployed grasps are successful. Success ratio is the highest on the FR3 robot. Note that average $\eta^{sim2real}$ obtained with Baxter is underestimated compared to UR5-SIH and FR3, as the longer experiment allowed to do more test on adversarial objects. Overall, this experiment showed that the \textbf{QD-generated grasps can successfully be transferred into the real world on a large diversity of setups}, including challenging objects (Fig. \ref{fig:diverse_grasps_s2r_franka}). It is worth noting that the optimization of grasps stability proposed with ME-scs in \cite{huber2023quality} makes the average success ratio reach $41\%$, an increase of $8\%$ compared to the diversity search methods (NSMBS and DC-NSMBS). Nevertheless, all the deployed trajectories were successful in the similar simulated scene, stressing the need to identify the causes of transferability issues, and find ways to alleviate their impact.


\textit{\textbf{Reality gap for grasping.}}
Several reasons for transferability failures can clearly be identified (Fig. \ref{fig:reality_gap_issues}): \textbf{1) Errors in frictions.} The simulated scene considers a unique set of friction coefficients for each objects, while their surface are usually composed of different textures. Fig. \ref{fig:reality_gap_issues}.a shows a successful grasp in simulation that slips from the finger in reality; \textbf{2) Coarse modelization of contacts.} Inaccurate modelization of collisions can lead to successful grasps that are infeasible in reality (e.g. insert the gripper in some part of the object to lift it, Fig. \ref{fig:reality_gap_issues}.b); \textbf{3) Errors in inertia.} A coarse approximation of the object distribution of mass can lead to unrealistic interaction dynamics, like applying forces on an object side to lift it and grab it in the air (Fig. \ref{fig:reality_gap_issues}.c);
\textbf{4) Coarse approximations of the 3D models.} The difference between a simulated model and the real object is often a cause of transfer failure. Fig. \ref{fig:reality_gap_issues}.d shows that the coarse model of YCB padlock results into grasps on inexisting parts of the object. Note that the less rigid objects can be alterated due to the multiple interactions with the grippers, resulting into a different shape; \textbf{5) Fragile grasps.} Some grasps found in deterministic simulation can simply be not robust enough to successfully transfer into the real world (Fig. \ref{fig:reality_gap_issues}.e). Other notable problems include \textbf{coarse modelization of actuators dynamics}, \textbf{joint and perception noise}, and \textbf{hardware setup constraints} (e.g. the SIH cable prevented some trajectories).

\textit{\textbf{Quality criteria.}}
Fig. \ref{fig:dr_fit_sim2real_transfert_ratio_results} shows the obtained results on any of the tested criteria. The displayed lines (right and center columns) correspond to the linear regression on the collected measures. There is a correlation between the sim2real transferability ratio $\eta^{sim2real}$ and the DR-based fitnesses, regardless of the platform. Overall, the average Pearson coefficient is $r=0.27 \pm 0.07$. The worse discrimative criterion is $f^{JSDR}$ with $r=0.24 \pm 0.09$, $p \leq 0.01$. It is worth noting that DR more reliably discriminates transferable grasps from others on the dexterous hand than on 2-fingers grippers. Such a difference might come from the wrong modelization of contacts, which tend to optimistically evaluate the success of antipodal grasps. On the left column of Fig. \ref{fig:dr_fit_sim2real_transfert_ratio_results} is shown the obtained measures on the robot under which the larger amount of trajectories have been deployed (Baxter), aggregated into $30$ bins. Interestingly, some variants lead to more uniformly distributed measures of DR fitness (e.g. Object state DR and Mixture DR). Others are less balanced: There are fewer trajectories with $f^{JSDR} \approx 0.5$ than closer to the boundaries. Even more extreme is Joint Poses DR, which associate trajectories with either $f^{JSDR} = 0$ or $f^{JSDR} = 1$ (bins with less than $30$ solutions are discarded). 

\textbf{These results validate the intuition that the proposed variants of DR are positively correlated with sim2real transferability.} Such quality criteria can be used to distinguish the most promising solutions from the exploitation of simulations. Furthermore, optimizing robustness to DR should lead to a higher sim2real transferability ratio.

\begin{figure}[t]
  \centering
  \includegraphics[width=\columnwidth]{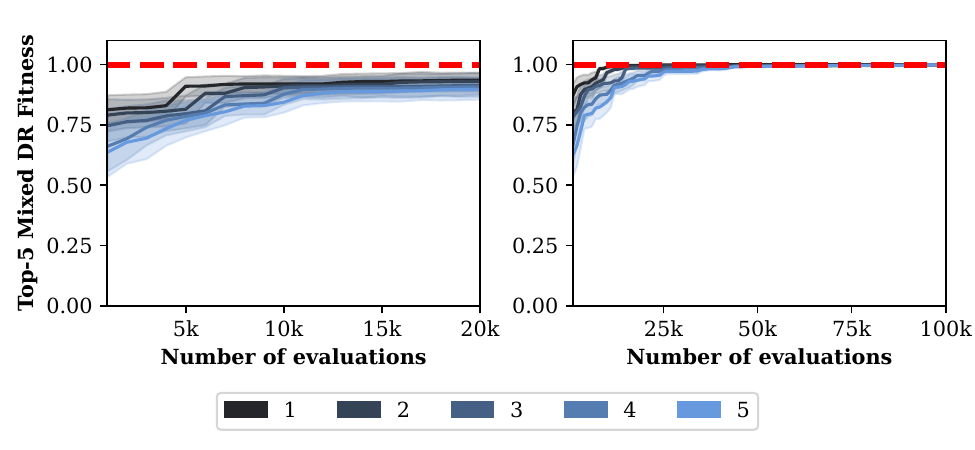}
  \caption{\textbf{TR-ME optimization of the Mixed Domain Randomization fitness on the FR3 robot.} The red dashed line is the optimal fitness. (Left) results on 20k evaluations are used for the experiments on the real robot. (Right) convergence in the longer run.}
  \label{fig:tr_me_top5_fits_optim}
\end{figure}

\textit{\textbf{Grasps robustification.}}
Fig. \ref{fig:tr_me_top5_fits_optim} shows the evolution of top-5 best-performing solutions throughout the TR-ME optimization process. Their performances tend to the optimal solution, meaning that all the simulated deployments result in a successful grasp for any of the sampled DR perturbations. Table \ref{table:s2r_reg_model_results} contains the sim2real transfer ratios for the tested variants on the FR3 arm. ME-scs ratios correspond to the results obtained in the systematic sim2real study. Selecting the most robust solutions to Mixture DR leads to an average ratio of $\eta^{sim2real}=0.60$. After the robustification step proposed in TR-ME, the transfer ratio of the best-performing solutions reaches $\eta^{sim2real}=0.84$. This shows that the DR correlation with the sim2real transferability can successfully be exploited to produce more robust grasps. An approach like TR-ME can thus be used to generate large datasets of diverse and transferable reach-and-grasp trajectories. Such datasets can be used to bootstrap closed-loop learning methods or be straightforwardly used with a vision-based approach for pseudo-closed-loop grasps into the real world \cite{helenon2023learning}.


\begin{table}[t]
\centering
\begin{tabular}{ ||c || c | c | c || c ||}
\hline
\textbf{methods} & \multicolumn{3}{c||}{\textbf{ME-scs}} & \textbf{TR-ME} \\
\hline
selection & smallest-5 fit  & random & top-5 fit & top-5 fit \\
\hline
 $\eta^{sim2real}$  & 0.38 & 0.44 & 0.60 & \textbf{0.84} \\
\hline
\end{tabular}
\caption{\textbf{Impact of TR-ME robustification on the sim2real transfer ratios (FR3 robot).}}
\label{table:s2r_reg_model_results}
\end{table}



\textit{\textbf{Limitations.}} Reproducing these experiments on other platforms, even on the same ones but with different wear-and-tear, could lead to different results. Plus, the variability of QD algorithms outcome – especially on grasping \cite{huber2023quality} – might affect new collected measures. This is the reason why the present work focused on the study of correlations, drawing conclusions from \textbf{emerging main patterns that can be observed on different robotic platforms}, including \textbf{grippers of different natures}, and grasps produced with \textbf{different QD methods}. Such a variety of experimental conditions is \textbf{likely to make the obtained results hold on different grasping setups.} 

It is also worth noting that most of the trajectories have been deployed on the Baxter robot. Nevertheless, similar correlations are observed on the UR5-SIH and the FR3. Further, the grasp robustification experiment conducted on the FR3 shows that the sim2real transfer ratio can significantly be increased by optimizing robustness to DR, confirming the discriminating power of the proposed quality criteria regarding the reality gap.

The proposed DR variants can be leveraged to select the most promising solutions from large grasping datasets. Another interesting usage of such criteria is their \textbf{discriminative power to identify the most fragile generated solutions.} The exploratory power of QD methods can be leveraged to generate solutions that exploit simulation weaknesses, and the DR quality criteria can later allow the roboticists to identify the main weaknesses of the used simulator. These weaknesses can therefore be addressed to increase the sim2real exploitability of automatically generated grasps. Here, several crucial issues have been identified, suggesting new ideas to robustify the solutions. Especially, a domain randomization on inertia, if well done, could be efficiently discriminative. But defining this DR variant to make it mimic the real dynamics is an open problem.


\section{CONCLUSIONS}

This work demonstrates the exploitability of Quality-Diversity generated reach-and-grasp trajectories through the deployment of about 7000 grasping trajectories on 3 different robotic platforms. The results collected on the real robots led to the empirical demonstration that variants of Domain Randomization can be leveraged to make the generated grasps more robust to sim2real transfer. Selecting the best-performing solutions using DR-based quality criteria increases the success ratio by $36\%$ on a real robot, and refining the solution with our proposed QD approach increases the success ratio by $91\%$. We believe this work to be a significant step
toward the exploitation of automatically generated datasets of grasping trajectories. Such datasets bring a solution to the hard exploration problem of grasping by providing a large diversity of demonstrations on many different robotic platforms with minor adaptations.


\section*{ACKNOWLEDGMENT}

This work was supported by the Sorbonne Center for Artificial Intelligence, the German Ministry of Education and Research (BMBF) (01IS21080), the French Agence Nationale de la Recherche (ANR) (ANR-21-FAI1-0004) (Learn2Grasp), the European Commission's Horizon Europe Framework Programme under grant No 101070381 and from the European Union's Horizon Europe Framework Programme under grant agreement No 101070596. This work used HPC resources from GENCI-IDRIS (Grant 20XX-AD011014320). Authors deeply thank Pr. Sven Behnke and the members of the AIS lab of Bonn for their warm welcome and support.





\begin{thebibliography}{99}

\bibitem{hodson2018gripping} Hodson, R. (2018). A gripping problem: designing machines that can grasp and manipulate objects with anything approaching human levels of dexterity is first on the to-do list for robotics. Nature.

\bibitem{nguyen1988constructing} Nguyen, V. D. (1988). Constructing force-closure grasps. The International Journal of Robotics Research, 7(3), 3-16.

\bibitem{zhang2022robotic} Zhang, H., Tang, J., Sun, S., Lan, X. (2022). Robotic grasping from classical to modern: A survey. arXiv preprint arXiv:2202.03631.

\bibitem{huber2023quality} Huber, J., Hélénon, F., Coninx, M., Ben Amar, F., Doncieux, S. (2023). Quality Diversity under Sparse Reward and Sparse Interaction: Application to Grasping in Robotics. arXiv:2308.05483

\bibitem{qin2022from} Qin, Y., Su, H., Wang, X. (2022). From one hand to multiple hands: Imitation learning for dexterous manipulation from single-camera teleoperation. IEEE Robotics and Automation Letters, 7(4), 10873-10881.

\bibitem{fang2020graspnet} Fang, H. S., Wang, C., Gou, M., Lu, C. (2020). Graspnet-1billion: A large-scale benchmark for general object grasping. In Proceedings of the IEEE/CVF conference on computer vision and pattern recognition (pp. 11444-11453).

\bibitem{yang2023pave} Yang, J., Tan, W., Jin, C., Liu, B., Fu, J., Song, R., Wang, L. (2023). Pave the Way to Grasp Anything: Transferring Foundation Models for Universal Pick-Place Robots. arXiv preprint arXiv:2306.05716.

\bibitem{deng2009imagenet} Deng, J., Dong, W., Socher, R., Li, L. J., Li, K., Fei-Fei, L. (2009, June). Imagenet: A large-scale hierarchical image database. In 2009 IEEE conference on computer vision and pattern recognition (pp. 248-255). Ieee.

\bibitem{ray2023chatgpt} Ray, P. P. (2023). ChatGPT: A comprehensive review on background, applications, key challenges, bias, ethics, limitations and future scope. Internet of Things and Cyber-Physical Systems.

\bibitem{godlfeder2009columbia} Goldfeder, C., Ciocarlie, M., Dang, H., Allen, P. K. (2009, May). The columbia grasp database. In 2009 IEEE international conference on robotics and automation (pp. 1710-1716). IEEE.

\bibitem{mahler2017dexnet2} Mahler, J., Liang, J., Niyaz, S., Laskey, M., Doan, R., Liu, X., ... Goldberg, K. (2017). Dex-net 2.0: Deep learning to plan robust grasps with synthetic point clouds and analytic grasp metrics. arXiv preprint arXiv:1703.09312.

\bibitem{depierre2018jacquard} Depierre, A., Dellandréa, E., Chen, L. (2018, October). Jacquard: A large scale dataset for robotic grasp detection. In 2018 IEEE/RSJ International Conference on Intelligent Robots and Systems (IROS) (pp. 3511-3516). IEEE.

\bibitem{eppner2021acronym} Eppner, C., Mousavian, A., Fox, D. (2021, May). Acronym: A large-scale grasp dataset based on simulation. In 2021 IEEE International Conference on Robotics and Automation (ICRA) (pp. 6222-6227). IEEE.

\bibitem{akkaya2019solving} Akkaya, I., Andrychowicz, M., Chociej, M., Litwin, M., McGrew, B., Petron, A., ... Zhang, L. (2019). Solving rubik's cube with a robot hand. arXiv preprint arXiv:1910.07113.

\bibitem{muratore2022robot} Muratore, F., Ramos, F., Turk, G., Yu, W., Gienger, M., Peters, J. (2022). Robot learning from randomized simulations: A review. Frontiers in Robotics and AI, 31.

\bibitem{jiang2011efficient} Jiang, Y., Moseson, S., Saxena, A. (2011, May). Efficient grasping from rgbd images: Learning using a new rectangle representation. In 2011 IEEE International conference on robotics and automation (pp. 3304-3311). IEEE 

\bibitem{dasari2019robonet} Dasari, S., Ebert, F., Tian, S., Nair, S., Bucher, B., Schmeckpeper, K., ... Finn, C. (2019). Robonet: Large-scale multi-robot learning. arXiv preprint arXiv:1910.11215.

\bibitem{turpin2023fast} Turpin, D., Zhong, T., Zhang, S., Zhu, G., Liu, J., Singh, R., ... Garg, A. (2023). Fast-Grasp'D: Dexterous Multi-finger Grasp Generation Through Differentiable Simulation. arXiv preprint arXiv:2306.08132.

\bibitem{miller2004graspit} Miller, A. T., Allen, P. K. (2004). Graspit! a versatile simulator for robotic grasping. IEEE Robotics \& Automation Magazine, 11(4), 110-122.

\bibitem{lundell2021multi} Lundell, J., Corona, E., Le, T. N., Verdoja, F., Weinzaepfel, P., Rogez, G., ... Kyrki, V. (2021, May). Multi-fingan: Generative coarse-to-fine sampling of multi-finger grasps. In 2021 IEEE International Conference on Robotics and Automation (ICRA) (pp. 4495-4501). IEEE.

\bibitem{siddiqui2021grasp} Siddiqui, M. S., Coppola, C., Solak, G., Jamone, L. (2021). Grasp stability prediction for a dexterous robotic hand combining depth vision and haptic bayesian exploration. Frontiers in Robotics and AI, 8, 703869.

\bibitem{koos2012transferability} Koos, S., Mouret, J. B., Doncieux, S. (2012). The transferability approach: Crossing the reality gap in evolutionary robotics. IEEE Transactions on Evolutionary Computation, 17(1), 122-145.

\bibitem{jimenez2021model} Jimenez-Vazquez, E., Ayala-Rodriguez, J., Navarro-Duran, D., Lopez-Caudana, E. (2021, January). Model approximation of an arm of the nao™ robot using system identification. In 2021 Second International Symposium on Instrumentation, Control, Artificial Intelligence, and Robotics (ICA-SYMP) (pp. 1-6). IEEE.

\bibitem{jiang2021simgan} Jiang, Y., Zhang, T., Ho, D., Bai, Y., Liu, C. K., Levine, S., Tan, J. (2021, May). Simgan: Hybrid simulator identification for domain adaptation via adversarial reinforcement learning. In 2021 IEEE International Conference on Robotics and Automation (ICRA) (pp. 2884-2890). IEEE.

\bibitem{kadian2020sim2real} Kadian, A., Truong, J., Gokaslan, A., Clegg, A., Wijmans, E., Lee, S., ... Batra, D. (2020). Sim2real predictivity: Does evaluation in simulation predict real-world performance?. IEEE Robotics and Automation Letters, 5(4), 6670-6677.

\bibitem{collins2019quantifying} Collins, J., Howard, D., Leitner, J. (2019, May). Quantifying the reality gap in robotic manipulation tasks. In 2019 International Conference on Robotics and Automation (ICRA) (pp. 6706-6712). IEEE.

\bibitem{enayati2023facilitating} Enayati, A. M. S., Dershan, R., Zhang, Z., Richert, D., Najjaran, H. (2023). Facilitating Sim-to-Real by Intrinsic Stochasticity of Real-Time Simulation in Reinforcement Learning for Robot Manipulation. IEEE Transactions on Artificial Intelligence.

\bibitem{cully2022quality} Cully, A., Mouret, J. B., Doncieux, S. (2022, July). Quality-diversity optimisation. In Proceedings of the Genetic and Evolutionary Computation Conference Companion (pp. 864-889).

\bibitem{morel2022automatic} Morel, A., Kunimoto, Y., Coninx, A., Doncieux, S. (2022, May). Automatic acquisition of a repertoire of diverse grasping trajectories through behavior shaping and novelty search. In 2022 International Conference on Robotics and Automation (ICRA) (pp. 755-761). IEEE.

\bibitem{huber2023diversity} Huber, J., Sane, O., Coninx, M., Ben Amar, F., Doncieux, S. (2023, July). Diversity Search for the Generation of Diverse Grasping Trajectories. In Proceedings of the Companion Conference on Genetic and Evolutionary Computation (pp. 151-154).

\bibitem{mouret2015illuminating} Mouret, J. B., Clune, J. (2015). Illuminating search spaces by mapping elites. arXiv preprint arXiv:1504.04909.

\bibitem{bai2020survey} Bai, Q., Li, S., Yang, J., Song, Q., Li, Z., Zhang, X. (2020). Object detection recognition and robot grasping based on machine learning: A survey. IEEE access, 8, 181855-181879.

\bibitem{asada2011special} H. H. Asada and V. Kumar, “Special issue on stochasticity in robotics and bio-systems,” pp. 503–504, 2011.

\bibitem{ibarz2021train} Ibarz, J., Tan, J., Finn, C., Kalakrishnan, M., Pastor, P., Levine, S. (2021). How to train your robot with deep reinforcement learning: lessons we have learned. The International Journal of Robotics Research, 40(4-5), 698-721.

\bibitem{collins2021review} Collins, J., Chand, S., Vanderkop, A., Howard, D. (2021). A review of physics simulators for robotic applications. IEEE Access, 9, 51416-51431.

\bibitem{mordotach2015ensemble} Igor Mordatch, Kendall Lowrey, and Emanuel Todorov. Ensemble-cio: Full-body dynamic motion planning that transfers to physical humanoids. In Intelligent Robots and Systems (IROS), 2015 IEEE/RSJ

\bibitem{antonova2017reinforcement} Antonova, R., Cruciani, S., Smith, C., Kragic, D. (2017). Reinforcement learning for pivoting task. arXiv preprint arXiv:1703.00472.

\bibitem{helenon2023learning} Hélénon, H., Huber, J., Ben Amar, F., Doncieux, S. (2023). Learning to Grasp: from Somewhere to Anywhere. arXiv preprint

\bibitem{calli2015benchmarking} Calli, B., Walsman, A., Singh, A., Srinivasa, S., Abbeel, P., Dollar, A. M. (2015). Benchmarking in manipulation research: The ycb object and model set and benchmarking protocols. arXiv preprint arXiv:1502.03143.

\bibitem{coumans2016pybullet} Coumans, E., Bai, Y. (2016). Pybullet, a python module for physics simulation for games, robotics and machine learning.

\bibitem{mouret2017reality} Mouret, J. B., Chatzilygeroudis, K. (2017, July). 20 years of reality gap: a few thoughts about simulators in evolutionary robotics. In Proceedings of the Genetic and Evolutionary Computation Conference Companion (pp. 1121-1124).

\bibitem{newbury2023review} Newbury, R., Gu, M., Chumbley, L., Mousavian, A., Eppner, C., Leitner, J., ... , Cosgun, A. (2023). Deep learning approaches to grasp synthesis: A review. IEEE Transactions on Robotics.

\bibitem{salvato2021crossing} Salvato, E., Fenu, G., Medvet, E., Pellegrino, F. A. (2021). Crossing the reality gap: A survey on sim-to-real transferability of robot controllers in reinforcement learning. IEEE Access, 9, 153171-153187.


\end{thebibliography}
\end{document}